%% file: CameraReady-8393.tex
\documentclass[letterpaper]{article} 
\usepackage{aaai24}  
\usepackage{times}  
\usepackage{helvet}  
\usepackage{courier}  
\usepackage[hyphens]{url}  
\usepackage{graphicx} 
\usepackage{natbib}  
\usepackage{caption} 
\usepackage{algorithm}
\usepackage{algorithmic}

\usepackage{amsmath}
\usepackage{epsfig}
\usepackage{graphicx}
\usepackage{amssymb}
\usepackage{fancyhdr}
\usepackage{booktabs}
\usepackage{multirow}
\usepackage[T1]{fontenc}
\usepackage{subfigure}

\usepackage{newfloat}
\usepackage{listings}
\DeclareCaptionStyle{ruled}{labelfont=normalfont,labelsep=colon,strut=off} 
\floatstyle{ruled}
\newfloat{listing}{tb}{lst}{}
\floatname{listing}{Listing}
\title{Bias-Conflict Sample Synthesis and Adversarial Removal Debias Strategy for Temporal Sentence Grounding in Video}

\author{
	Zhaobo Qi\textsuperscript{\rm 1}, 
	Yibo Yuan\textsuperscript{\rm 1}, 
	Xiaowen Ruan\textsuperscript{\rm 1}, 
	Shuhui Wang\textsuperscript{\rm 2}, \\
	Weigang Zhang\textsuperscript{\rm 1}\thanks{Corresponding author.}, 
	Qingming Huang\textsuperscript{\rm 2, \rm 3 *}
}
\affiliations {
	\textsuperscript{\rm 1}Harbin Institute of Technology, Weihai, China\\
	\textsuperscript{\rm 2}Key Lab of Intell. Info. Process., Inst. of Comput. Tech., CAS, Beijing, China\\
	\textsuperscript{\rm 3}University of Chinese Academy of Sciences, Beijing, China\\
	qizb@hit.edu.cn, \{yuanyibo21,22s130489\}@stu.hit.edu.cn, wangshuhui@ict.ac.cn, \\ wgzhang@hit.edu.cn, qmhuang@ucas.ac.cn
}

\usepackage{bibentry}

\begin{document}

\maketitle

\begin{abstract}
	Temporal Sentence Grounding in Video (TSGV) is troubled by dataset bias issue, which is caused by the uneven temporal distribution of the target moments for samples with similar semantic components in input videos or query texts. Existing methods resort to utilizing prior knowledge about bias to artificially break this uneven distribution, which only removes a limited amount of significant language biases. In this work, we propose the bias-conflict sample synthesis and adversarial removal debias strategy (BSSARD), which dynamically generates bias-conflict samples by explicitly leveraging potentially spurious correlations between single-modality features and the temporal position of the target moments. Through adversarial training, its bias generators continuously introduce biases and generate bias-conflict samples to deceive its grounding model. Meanwhile, the grounding model continuously eliminates the introduced biases, which requires it to model multi-modality alignment information. BSSARD will cover most kinds of coupling relationships and disrupt language and visual biases simultaneously. Extensive experiments on Charades-CD and ActivityNet-CD demonstrate the promising debiasing capability of BSSARD. Source codes are available at~\url{https://github.com/qzhb/BSSARD}.
\end{abstract}

\input{Introduction}

\input{RelatedWork}

\input{Method}

\input{Experiment}

\section{Conclusion}
In this paper, we propose a novel adversarial training debias framework for TSGV task. 
It dynamically generates bias-conflict samples by exploiting all kinds of spurious relationships between single-modality features and the target moments, which can effectively remove the bias dependency of the TSGV models. 
In essence, the debias is achieved through directional data augmentation, which breaks the uneven temporal distribution of the target moments for samples that contain similar semantic components.  
Extensive experiments on Charades-CD and ActivityNet-CD datasets demonstrate the effectiveness of our strategy. 
In the future, we will apply our method to a wider range of scenarios, such as debias in visual question-answering task, and also eliminate cross-modality combination bias problems.

\section{Acknowledgments}
This work was supported in part by the National Natural Science Foundation of China under Grant U21B2038, 62306092, 61976069, 62022083, and 62236008.

\bibliography{aaai24}

\end{document}

%% file: Introduction.tex
\section{Introduction}
\label{sec:intro}

Temporal sentence grounding in video (TSGV) aims to locate the target video clip corresponding to the query sentence, which plays a vital role in fields such as video creation and video monitoring. 
With the help of deep learning, TSGV has achieved promising performance~\cite{RN2, RN3, RN4, RN5, zheng2023generating, wang2023scene}. 
However, recent studies~\cite{RN7, RN28, RN29} have shown that TSGV models are susceptible to dataset biases. 
They perform the grounding task through shortcuts created by dataset biases rather than the alignment between query texts and corresponding visual regions, resulting in poor generalization performance.

\begin{figure}[t]
	\centering
	\includegraphics[width=\linewidth]{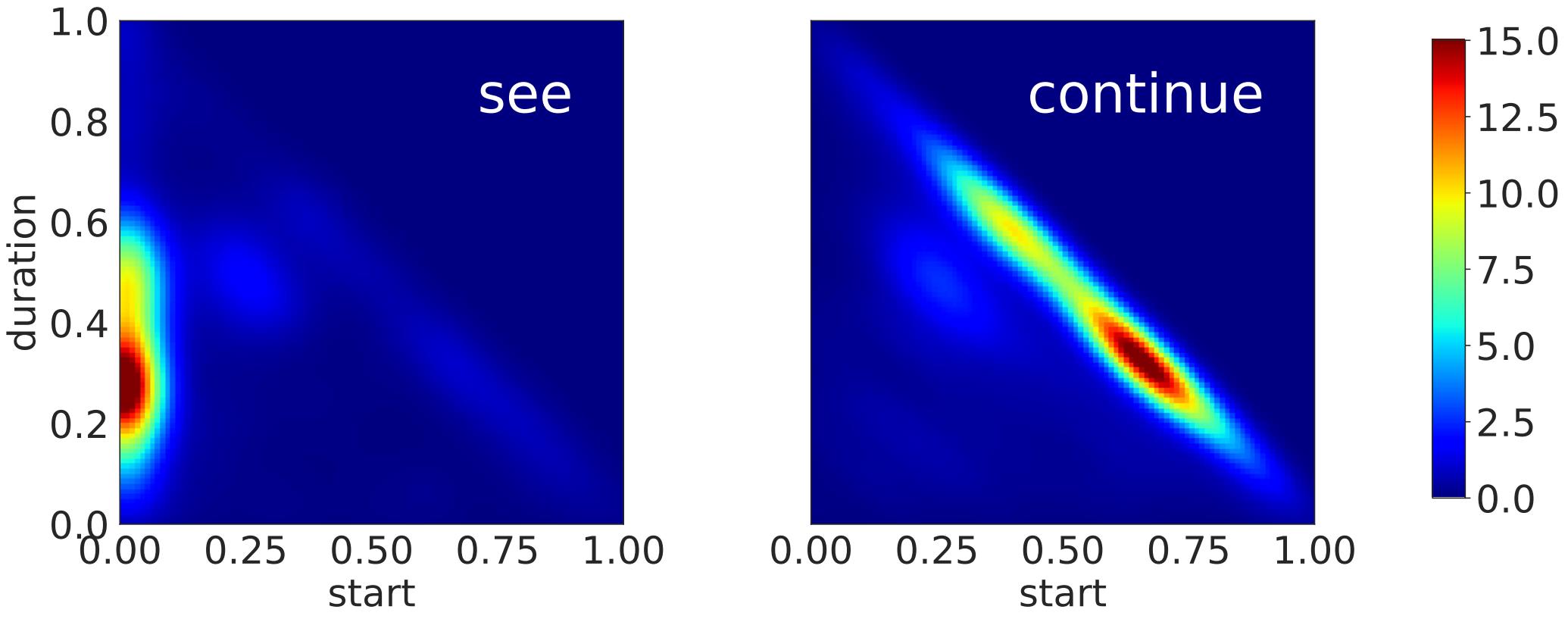}
	\caption{The spurious correlation between query words and temporal location of target moments. The horizontal and vertical axes represent the normalized starting time and duration of the target moment, respectively. The color represents the text-video pair density, which is obtained by using kernel density estimation with the Gaussian kernel.}
	\label{Introduction}
\end{figure}

From the perspective of the dataset distribution, 
given samples that contain similar semantic components in the input video or query text, the bias originates from the uneven temporal distribution of the corresponding target moments. 
As shown in Figure~\ref{Introduction}, in the widely used ActivityNet dataset, for samples containing the word ``see'' in the query text, their target moments tend to be concentrated in the first half of the video, and only a few target moments are located elsewhere in the video. This uneven distribution leads to unexpected spurious correlations between query words and target moments, which are easily captured by TSGV models. Once the query contains the word ``see'', the model will predict the target moment as the first half of the video with a high probability. 
Similar phenomena also exist in the visual modality. 
Hence, generating samples to break these correlations and form a uniform temporal distribution of target moments is an effective debiasing strategy.

However, due to the rich semantic content in the video and text, we can not enumerate all the coupling relationships between the target moment and the semantic component of the video/text to generate corresponding samples. It is impossible to construct an ideal dataset without dataset biases. 
Therefore, Zhang {\it et al.}~\cite{RN31} modify the relative positions of target moments through video clipping to form a more uniform temporal distribution of target moments. 
Liu {\it et al.}~\cite{RN28} replace certain verbs or nouns in the query with similar vocabulary from the dataset to obtain negative samples for the coupling relationship between query words and target moments. 
These methods only disrupt a fixed quantity of dataset biases because of the limited diversity of words in the dataset and the limited relative positions of target moments. 

Based on the above analysis, we propose the bias-conflict sample synthesis and adversarial removal debias strategy (BSSARD) for TSGV. 
Its core idea is that in the training stage, we randomly generate fake target moments for each real sample. Then we use the bias generator to generate visual or linguistic bias features condition on these fake target moments and construct bias-conflict samples that are falsely associated with the synthetic target moments. Through adversarial training, the bias discriminator should accurately judge that the bias-conflict sample contains biases and predict the original target moment. 
In essence, it will generate some samples for all the spurious correlations between the temporal position of the target moments and the visual/text modality data. 
It greatly destroys the uneven temporal distribution of the target moments in the TSGV datasets and disrupts language and visual biases simultaneously. 

Specifically, BSSARD consists of two bias generators and a grounding model with a bias discriminator. Given a video-text pair $(I_v,I_q)$ with target moment $(y_s^r, y_e^r)$, BSSARD will iteratively generate visual/text bias-conflict samples to form real and fake samples for adversarial debiasing. We first randomly generate a fake target moment $(y^f_s, y^f_e)$. 
Mimicking the spurious correlation between the visual feature and the fake target moment, we use the visual bias generator to produce a bias vector based on $(y^f_s, y^f_e)$, and add it to the visual feature of $I_v$, which constructs the bias-conflict sample $(\tilde{I_v},I_q)$. Through $(\tilde{I_v},I_q)$, the visual bias generator aims to force the grounding model to produce $(y^f_s, y^f_e)$ and determine the input is a sample without bias. 
Instead, the grounding model needs to produce $(y_s^r, y_e^r)$ for $(I_v,I_q)$ and determine it is a sample without bias. Meanwhile, it should also produce $(y_s^r, y_e^r)$ for $(\tilde{I_v},I_q)$ and determine it is a sample with bias. 
For the query bias generator, the process is similar. 
Through adversarial training, bias generators continuously introduce biases and generate bias-conflicting samples to deceive the grounding model.
Meanwhile, the grounding model continuously eliminates the introduced biases, which requires it to model cross-modality alignment information to accomplish the grounding task. 
Hence, the final model can effectively resist the influence of dataset biases and achieve better generalization performance.

Experimental results on the repartitioned Charades-CD and ActivityNet-CD datasets demonstrate the effective debiasing capability of our approach. 

Our main contributions are as follows: 
\begin{itemize}
	\item We propose a novel debiasing strategy, which generates bias-adversarial samples for most dataset biases and performs debias through adversarial training. 
	\item We propose a debias network with two bias generators, a grounding block, and a bias discriminator, which can effectively eliminate visual and language biases. 
	\item Extensive experiments on the OOD-test set demonstrate that BSSARD achieves great debias performance.
\end{itemize}

%% file: RelatedWork.tex
\section{Related Work}

\subsection{Temporal Sentence Grounding in Video}

TSGV methods are generally divided into proposal-based and proposal-free models.
Proposal-based methods first generate a large number of candidate proposals, and then score all proposals with the language queries. 
Anchor-based proposal generation methods~\cite{RN8, RN9, RN10, RN11, RN5} pre-define some boxes of specific proportions and directly generate proposals on the multi-modal features. 2D-based proposal generation methods~\cite{RN2, RN13, RN14, RN15, RN16, RN17} extend anchors to 2D to more finely model the positional relationships between proposals.
Proposal-free methods are divided into regression-based and span-based methods. Regression-based methods~\cite{RN18, RN19, RN20, RN21, RN3, zheng2023generating, wang2023scene} directly regress start and end timestamps. Such methods mainly focus on designing various feature encoding and cross-modal reasoning strategies to achieve precise localization. Span-based methods~\cite{RN22, RN23, RN25, RN26, RN4, zheng2023phrase} directly predict the probability of each video segment as the start and end position of the target moment.

\subsection{Model Debiasing}

Solutions for mitigating dataset bias are divided into two categories: ensemble-based methods and additional constraint-based methods.
Ensemble-based methods~\cite{RN34, RN36, RN28} involve using weak models to learn the bias and then assessing the bias of samples based on the weak models. Biased samples are then given lower weights in the main model to achieve debias. 
Constraint-based methods~\cite{RN30, RN35, RN29, lan2023curriculum, qi2023self, lan2022closer, yoon2023counterfactual} usually entail designing special structures and adding extra tasks to the model. This approach can separate bias from true rules, forcing the model to learn the true rules to complete the task. 
In contrast to the above methods, we propose to synthesize bias-conflict samples and disrupt the uneven distribution of the target moments for samples with similar semantic components, which forces models to learn useful alignment visual-language information for grounding.


%% file: Method.tex
\begin{figure*}[t]
	\centering
	\includegraphics[width=0.96\linewidth]{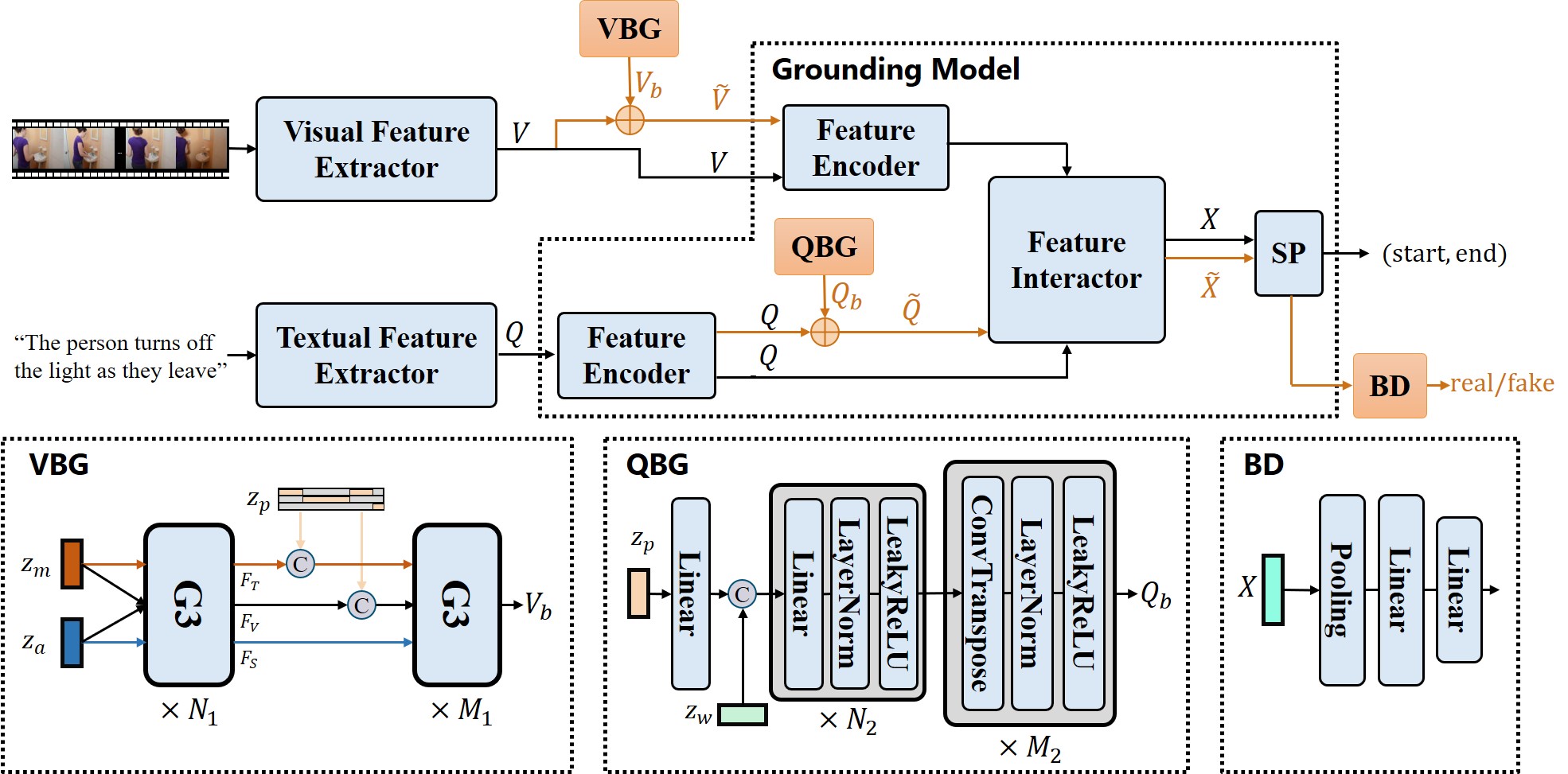}
	\caption{An overview of our BSSARD. The orange background is only used during the training period. Best viewed in color.}
	\label{Overview}
\end{figure*}

\section{Method}

\subsection{Overview}

Our bias-conflict sample synthesis and adversarial removal debias strategy (BSSARD) is shown in Figure~\ref{Overview}, which is based on a span-based grounding model and incorporates a Visual Bias Generator (VBG), a Query Bias Generator (QBG), and a Bias Discriminator (BD). 

In the training stage, given a video-query pair~$(I_v,I_q)$ with the target moment~$(y_s^r,y_e^r)$ , we first utilize the visual and textual feature extractor to capture feature representations $V$ and $Q$. 
On the one hand, we feed the original $V$ and $Q$ into the feature encoder and feature interactor to obtain real sample feature $X$. Then, the span predictor (SP) should accurately produce the start and end time of the target moment. Meanwhile, the bias discriminator module should accurately predict it as the real sample. 
On the other hand, given a fake temporal location $z_p=(y_s^f, y_e^f)$ of the target moment obtained through random uniform sampling from the video frames, we employ VBG to generate visual bias feature $V_b$ and element-wise add it to the real sample feature $V$, which produces $\tilde{V}$. Similar to the real sample, we obtain bias-conflict sample feature $\tilde{X}$. It has two labels, one is the temporal position $(y_s^r,y_e^r)$ of the real target moment and the other is the fake temporal position $(y_s^f, y_e^f)$. The bias generator wants to generate shortcut features to trick SP into generating a target moment similar to $z_p$ and trick BD into judging the current sample as the real sample. Instead, the SP will try to generate target moments as close as $(y_s^r,y_e^r)$ and the BD will judge it as a sample with bias. 
Similarly, we also employ QBG to generate textual bias feature $Q_b$ and perform the above process. 
The synthesized bias features simulate all kinds of spurious correlations between the semantic information of video/text and the temporal position of the target clips. Through adversarial training, it effectively mitigates visual and linguistic biases. 
In the inference stage, the VBG, QBG, and BD components are removed.

\subsection{Bias Generator}

The bias generator generates bias-conflict sample features based on synthesized spurious relationships between the unimodality information and the temporal positional labels.

\subsubsection{Visual Bias Generator} 

We construct VBG by the $\text{G}^\text{3}$ module~\cite{G3AN}, which encodes video features by decoupling appearance and motion content. The VBG contains three randomly generated variables $z_a$, $z_m$, and $z_p$ as inputs, and generates visual bias vector $V_{b}$ for constructing bias-conflict samples. These random variables guarantee the diversity and randomness of the bias generator.

Especially, the $z_a$ and $z_m$ are random vectors that follow a normal distribution and represent appearance and motion content, respectively. 
VBG first inputs $z_a$ and $z_m$ into $N_1$ stacked $\text{G}^\text{3}$, and produces three visual stream features, named temporal stream $F_T$, spatial stream $F_S$ and video stream $F_V$. 
Then, we random generate a fake temporal label $z_p \in \mathbb{R}^{3 \times n}$ of the target clip, which is obtained through random uniform sampling from the video frame sequence within the range of $\left[0, n-1\right]$. Its three dimensions are mutually exclusive and represent whether the segment contains background, foreground, or ignored data. It provides the temporal position conditional information that the bias generator requires. Subsequently, $z_p$ is concatenated with $F_T$ and $F_V$ to introduce temporal position information. 
Next, VBG outputs the visual bias vector $V_b\in \mathbb{R}^{n \times d}$ through $M_1$ stacked $\text{G}^\text{3}$. Finally, we add $V_b$ to $V$ and obtain visual bias conflict sample feature $\tilde{V}$,
\begin{equation}
	\begin{split}
		V_{b} &= VBG\left(z_a, z_m, z_p\right)\\
		\tilde{V} &= V_b + V
	\end{split}	
	\label{eq1}
\end{equation}
where $n$ and $d$ are the video length and feature dimension.

\subsubsection{Query Bias Generator}

We construct QBG by a simple fully connected structure and transpose convolutional structure. 
The QBG needs two randomly generated variables $z_w$ and $z_p$ as inputs and generates textual bias vector $Q_{b}$ for constructing bias-conflict samples. 

Specifically, we randomly sample $z_p$ from a normal distribution, which is a random vector of length $n$ and represents the synthetic label of the fake target video moment. Then we embed it using a fully connected layer, which provides the temporal position information for the feature generation. 
To enhance randomness and diversity, QBG also generates a random vector $z_w$ from a normal distribution to introduce the context for the feature generation, which is concatenated with the encoded position feature. 
We use $N_2$ linear layers and $M_2$ transposed convolution layers to produce the word sequence bias feature $Q_b\in \mathbb{R}^{m \times d}$. Finally, we add $Q_b$ to $Q$ and obtain query bias conflict sample feature $\tilde{Q}$,
\begin{equation}
	\begin{split}
		Q_b &= QBG\left(z_w, z_p\right)\\ 
		\tilde{Q} &= Q_b + Q
	\end{split}
	\label{eq2}
\end{equation}
where $m$ represents the length of the query text.

\subsection{Grounding Model}

The grounding model consists of a feature encoder, feature interactor, span predictor (SP), and bias discriminator (BD). 
BD is a binary classifier that identifies whether the sample contains biases. 
SP should make correct temporal grounding predictions for both regular and bias-conflict samples. 

For the normal samples without synthetic bias features, $V$ and $Q$ are encoded by the feature encoder and feature interactors to generate cross-modal representations $X$. Subsequently, the SP and BD make predictions, 
\begin{equation}
	\begin{split}
		p_{s}^r, p_{e}^r &= SP(X) \in \mathbb{R}^{n}, \mathbb{R}^{n}\\
		p_{d}^r &= BD(X) \in \mathbb{R}^{2}
	\end{split}
	\label{eq3}
\end{equation}
where $p_{s}^r$ and $p_{e}^r$ represent the predicted start and end positions of the target moment for the real sample, and $p_{d}^r$ represents the probability of the sample containing bias.

For bias-conflict samples, we iteratively encode $\tilde{V}$ and $Q$ or $V$ and $\tilde{Q}$ in each training iteration by the feature encoder and feature interactor to generate cross-modal representation $\tilde{X}$. Subsequently, the SP and BD make predictions,
\begin{equation}
	\begin{split}
		p_{s}^f, p_{e}^f &= SP(\tilde{X}) \in \mathbb{R}^{n}, \mathbb{R}^{n}\\
		p_{d}^f &= BD(\tilde{X}) \in \mathbb{R}^{2}
	\end{split}
	\label{eq4}
\end{equation}
where $p_{s}^f$ and $p_{e}^f$ represent the predicted start and end positions of the target moment for the bias-conflict sample, and $p_{d}^f$ represents the probability of the sample containing bias.

\subsection{Training Objectives}

\subsubsection{Training Process}

As shown in Algorithm~\ref{Training Process}, we train the two bias generators alternately in each training step. We detail the bias generator loss $L^g$ and the discriminator loss $L^d$.

\begin{algorithm}[t]
	\caption{Training process in one epoch.}
	\label{Training Process}
	\begin{algorithmic}[1] 
		\REQUIRE $\text{VBG}$; $\text{QBG}$; $\text{GD}$;
		\ENSURE TSGV model
		\FOR{each training iteration}
		\STATE Sample $\left(V, Q \right)$, $(y_s^r,y_e^r)$
		\STATE Sample $z_m$, $z_a$, $z_p$, $(y_s^f,y_e^f)$
		\STATE $V_b \leftarrow \text{VBG}(z_m, z_a, z_p)$
		\STATE $\tilde{V} \leftarrow V + V_b$, $\tilde{Q} \leftarrow Q$
		\STATE Get $p_{s}^r, p_{e}^r, p_{d}^r, p_{s}^f, p_{e}^f, p_{d}^f$ by Equation~\ref{eq3} and~\ref{eq4} 
		\STATE Calculate $L^g$ and train VBG, freeze GD
		\STATE Calculate $L^d$ and train GD, freeze VBG
		
		\STATE Sample $z_w$, $z_p$, $(y_s^f,y_e^f)$
		\STATE $Q_b \leftarrow \text{QBG}(z_w, z_p)$
		\STATE $\tilde{V} \leftarrow V$, $\tilde{Q} \leftarrow Q + Q_b$
		\STATE Get $p_{s}^r, p_{e}^r, p_{d}^r, p_{s}^f, p_{e}^f, p_{d}^f$ by Equation~\ref{eq3} and~\ref{eq4} 
		\STATE Calculate $L^g$ and train QBG, freeze GD
		\STATE Calculate $L^d$ and train GD, freeze QBG
		\ENDFOR
	\end{algorithmic}
\end{algorithm}

\subsubsection{Bias Generator Training Objectives}
The visual/query bias generator aims to deceive the bias discriminator and induce the span predictor to produce the given fake labels.

We use cross-entropy loss to trick the bias discriminator,
\begin{equation}
	L_{cls}^g=f_{CE}(p_{d}^f, 0)
	\label{g_loss_cls}
\end{equation}
where $p_d^f$ represents the predicted sample flag. $0$ indicates the current sample is a real sample without synthetic bias. It encourages the generator to generate samples that the BD will classify as unbiased. 

To induce span predictor, we use cross-entropy loss:
\begin{equation}
	L_{loc}^g=\frac{1}{2}\left[f_{CE}\left(p_{s}^{f}, y_{s}^{f}\right)+f_{CE}\left(p_{e}^{f}, y_{e}^{f}\right)\right]
	\label{g_loss_loc}
\end{equation}
This encourages the bias generator to generate bias-conflict samples that can deceive the grounding model.

The total training loss of the bias generator is
\begin{equation}
	L^{g}=L_{loc}^g+\lambda_{1} L_{cls}^g
	\label{g_loss}
\end{equation}
where $\lambda_{1}$ is a weight hyperparameter.

\subsubsection{Discriminator Training Objectives}
The discriminator needs to distinguish between real and bias-conflict samples and predict the true target moments.

For the bias discriminator, we use cross-entropy loss,
\begin{equation}
	L_{cls}^d=f_{CE}(p_{d}^r, 0) + f_{CE}(p_{d}^f, 1)
	\label{d_loss_cls}
\end{equation}
It encourages the discriminator to accurately judge real and synthetic samples.

For the temporal grounding, we use the traditional span-based cross-entropy loss,
\begin{equation}
	\begin{split}
		L_{loc}^d =&\frac{1}{2}\left[f_{CE}\left(p_{s}^{r}, y_{s}^{r}\right)+f_{CE}\left(p_{e}^{r}, y_{e}^{r}\right)\right]\\
		+& \frac{1}{2}\left[f_{CE}\left(p_{s}^{f}, y_{s}^{r}\right)+f_{CE}\left(p_{e}^{f}, y_{e}^{r}\right)\right]\\
	\end{split}
	\label{d_loss_loc}
\end{equation}

Besides, we find it advantageous to incorporate a sample-based distance metric such as KL divergence. Consequently, we introduce an additional objective, 
\begin{equation}
	L^d_{kl} = D_{kl}(p_{s}^{r} \| p_{s}^{f}) + D_{kl}(p_{e}^{r} \| p_{e}^{f})
	\label{loss_kl_d}
\end{equation}
Here, Equation~\ref{d_loss_loc} aligns the predictions towards the real labels, and Equation~\ref{loss_kl_d} limits the predictions of bias-conflict samples to not deviate much from the predictions of the corresponding original samples. 
Meanwhile, KL divergence is a regularization method to prevent the generator from overfitting training data and improve the generalization ability.

\begin{table*}[t]
	\centering
	\renewcommand{\arraystretch}{1}
	\setlength{\tabcolsep}{1.4mm}
	\begin{tabular}{c c c c | c c c|c c c | c c c}
		\toprule
		\multirow{3}{*}{Method}
		& \multicolumn{6}{c}{Charades-CD} & \multicolumn{6}{c}{ActivityNet-CD} \\
		& \multicolumn{3}{c}{OOD} & \multicolumn{3}{c}{IID} & \multicolumn{3}{c}{OOD} & \multicolumn{3}{c}{IID}\\
		& r1i5 & r1i7 & mIoU & r1i5 & r1i7 & mIoU & r1i5 & r1i7 & mIoU & r1i5 & r1i7 & mIoU \\
		\midrule
		2D-TAN~\cite{RN2} & 28.18 & 13.73 & 34.22 & 46.48 & 28.76 & 42.73 & 18.86 & 9.77 & 28.31 & 40.87 & 28.95 & 44.99\\
		
		LG~\cite{RN21} & 42.90 & 19.29 & 39.43 & 51.28 & 28.68 & 45.16 &
		23.85 & 10.96 & 28.46 & 46.41 & 29.28 & 44.62\\
		
		DRN~\cite{RN20} & 31.11 & 15.17 & 23.05 & 42.04 & 23.32 & 28.21 & 
		- & - & - & - & - & -\\
		
		VSLNet~\cite{RN4} & 34.10 & 17.87 & 36.34 & 43.26 & 28.43 & 42.92 &
		20.03 & 10.29 & 28.18 & 47.81 & 29.07 & 46.33\\
		
		DCM~\cite{RN29} & 40.51 & 21.02 & 40.99 & 52.50 & 35.28 & 48.74 & 20.86 & 11.07 & 28.08 & 42.15 & 29.69 & 45.20\\
		
		SVTG~\cite{RN30} & 46.67 & 27.08 & 44.30 & 57.59 & 37.79 & 50.93 &
		24.57 & 13.21 & 30.45 & 48.07 & 32.15 & 47.03\\
		
		MDD~\cite{lan2022closer} & 40.39 & 22.70 & - & 52.78 & 34.71 & - & 20.80 & 11.66 & - & 43.63 & 31.44 & - \\
		
		\midrule
		QAVE~\cite{hao2022query} & 37.84 & 19.67 & 38.45 & 47.63 & 29.28 & 44.17 &
		21.39 & 10.86 & 28.41 & 44.58 & 27.42 & 44.28 \\
		
		BSSARD-QAVE & \bf 44.47 & \bf 26.16 & \bf 42.95 & \bf 54.31 & \bf 37.67 & \bf 49.41 & \bf 23.11 & \bf 12.17 &	\bf 29.40 & \bf 46.04 & \bf 30.06 & \bf 45.56 \\
		
		\midrule
		EXCL~\cite{RN3} & 39.61 & 19.35 & 38.81 & 47.18 & 26.59 & 44.02 & 23.38 & 12.66 &	29.19 & 45.76 & 29.62 & \bf 46.42 \\
		BSSARD-EXCL &  {\bf 41.93} & {\bf 22.53} & {\bf 40.9} & \bf 50.00 & \bf 31.86 & \bf 46.47 & \bf 23.75 & \bf 12.93 & \bf 29.26 & \bf 46.98 & \bf 30.34 & 46.12 \\
		
		\midrule
		VSLNet*~\cite{RN4} & 43.08 & 22.52 & 41.52 & 52.86 & 34.87 &  48.23 &
		25.40 & 13.51 & 30.33 & 48.07 & 31.19 & 46.94\\
		BSSARD-VSLNet* & {\bf 47.20} & {\bf 27.17} & {\bf 44.59} & {\bf 55.65} & {\bf 36.33} & {\bf 50.45} &
		{\bf 27.02} & {\bf 14.93} & {\bf 31.49} & {\bf 49.67} & {\bf 33.72} & {\bf 48.54}\\
		
		\bottomrule
	\end{tabular}
	\caption{Comparison results with state-of-the-arts. VSLNet*	replaces the encoder in VSLNet with a transformer block.}
	\label{sota-redivided}
\end{table*}

The total training loss of the discriminator is:
\begin{equation}
	L^d=L_{loc}^d+\lambda_{2} L_{cls}^d + \lambda_{3} L_{kl}^d
	\label{eq12}
\end{equation}
where $\lambda_{2}$ and $\lambda_{3}$ are weight hyperparameters.

%% file: Experiment.tex
\section{Experiment}

\subsection{Experiment Setup}

\subsubsection{Datasets}

We conduct experiments on the repartitioned Charades-CD and ActivityNet-CD~\cite{lan2022closer}. 
In these repartitioned datasets, the target moments of samples in the training, val, and test-iid sets are independent and identically distributed, while the test-ood set contains out-of-distribution samples. 
We can verify the debias ability of different models according to the grounding performance on the test-ood set and the performance difference between test-iid and test-ood sets.

\subsubsection{Metrics}
We use evaluation metrics of R@$n$, IoU=$m$, and mIoU. 
R@$n$, IoU=$m$ measures the proportion of test samples in which at least one of the top-$n$ localization results has an IoU score greater than $m$. 
The mIoU measures the average IoU score across all test samples. 
We set $n$ to 1 and $m$ to either 0.5 or 0.7, and use r1i5 and r1i7 to denote R@1, IoU=0.5 and R@1, IoU=0.7, respectively.

\subsubsection{Implementation Details}
For the text feature extractor, we use 300d GloVe~\cite{Glove} vectors for initialization. For the visual feature extractor, we use pre-trained I3D~\cite{I3D} and C3D~\cite{C3D} features. For the visual bias generator, we set $N_1$ to 4 and $M_1$ to 2. For the query bias generator, we set $N_2$ to 2 and $M_2$ to 4. During the training process, we use the AdamW~\cite{AdamW} optimizer, with a batch size of 16, and an initial learning rate of 0.001. We set $\lambda_{1}$ to 1, $\lambda_{2}$ to 1, and $\lambda_{3}$ to 1.

\begin{table}[t]
	\centering
	\renewcommand{\arraystretch}{1}
	\setlength{\tabcolsep}{1mm}
	\begin{tabular}{c|cc|ccc|ccc}
		\toprule
		\multirow{2}{*}{Row}
		& \multicolumn{2}{c}{BG} & \multicolumn{3}{c}{Grounding} &
		\multicolumn{3}{c}{OOD}  \\
		& $L_{loc}^g$ & $L_{cls}^g$ & $L_{loc}^d$ & $L_{cls}^d$ & $L_{kl}^d$
		& r1i5 & r1i7 & mIoU \\
		\midrule
		1 & - & - & $\checkmark$ & - & - &
		43.08 & 22.52 & 41.52 \\
		2 & $\checkmark$ & - & $\checkmark$ & - & - &
		46.64 & 26.40 & 43.49 \\
		3 & $\checkmark$ & - & $\checkmark$ & - & $\checkmark$ &
		45.99 & 26.16 & 44.02 \\
		4 & - & $\checkmark$ & $\checkmark$ & $\checkmark$ & - &
		45.16 & 25.72 & 43.18 \\
		5 & - & $\checkmark$ & $\checkmark$ & $\checkmark$ & $\checkmark$ & 
		46.22 & 26.40 & 44.22 \\
		6 & $\checkmark$ & $\checkmark$ & $\checkmark$ & $\checkmark$ & - & 
		{\bf 47.41} & 26.84 & 44.00 \\
		7 & $\checkmark$ & $\checkmark$ & $\checkmark$ & $\checkmark$ & $\checkmark$ &
		47.20 & {\bf 27.17} & {\bf 44.59} \\
		\bottomrule
	\end{tabular}
	\caption{Ablation study about loss functions.}
	\label{loss term study}
\end{table}

\subsection{Comparison with State-of-the-Arts}

To verify the universality of our debias strategy, we implement our BSSARD on multiple existing backbones QAVE, ExCL, and VSLNet*. The experiment results on Charades-CD and ActivityNet-CD datasets are presented in Table~\ref{sota-redivided}. 
We can see that our approach can significantly improve the grounding performance of different backbones on most evaluation metrics for OOD and IID test sets in both datasets. This indicates that our method has stronger debias and grounding ability. 
Besides, although our method shows the best performance in ActivityNet-CD's IID (BSSARD-VSLNet*), it works weaker than SVTG in Charades-CD's IID test set. This is due to the IID test set of Charades-CD containing a higher proportion of biased samples. 

\subsection{Ablation Study}

We conduct abundant ablation studies on Charades-CD datasets over BSSARD-VSLNet* backbones. 

\subsubsection{Loss Terms}

We analyze the impact of each loss function and their combinations on the grounding performance. Table~\ref{loss term study} summarizes the results. 
We can see that each loss function has its validity, and the combination of generation and adversarial losses can further improve the grounding performance on the OOD test set. 
Besides, the $L_{kl}^d$ leads to performance improvements in most cases. However, the improvement is limited when $L_{cls}^d$ is not used. This is because $L_{kl}^d$ depends on the grounding model's attention to the bias features brought by $L_{cls}^d$.

\begin{table}[t]
	\centering
	\renewcommand{\arraystretch}{1}
	\setlength{\tabcolsep}{0.4mm}
	\begin{tabular}{c c c c | c c c}
		\toprule
		\multirow{2}{*}{Method} & \multicolumn{3}{c}{OOD} & \multicolumn{3}{c}{IID}\\
		& r1i5 & r1i7 & mIoU & r1i5 & r1i7 & mIoU \\
		\midrule
		QAVE & 37.84 & 19.67 & 38.45 & 47.63 & 29.28 & 44.17 \\
		QAVE-Q & 41.96 & 23.20 & 40.97 & 50.18 & 32.56 & 47.11 \\
		QAVE-V & 42.79 & 22.34 & 41.03 & 54.07 & 33.41 & 48.31 \\
		BSSARD-QAVE & \bf 44.47 & \bf 26.16 & \bf 42.95 & \bf 54.31 & \bf 37.67 & \bf 49.41 \\
		\midrule
		ExCL & 39.61 & 19.35 & 38.81 & 47.18 & 26.59 & 44.02\\
		ExCL-Q & 39.94 & 19.7 & 39.05 & 48.90 & 26.59 & 44.72 \\
		ExCL-V & 39.10 & 19.21 & 38.77 & 47.79 & 30.76 &  45.58 \\
		BSSARD-ExCL & {\bf 41.93} & {\bf 22.53} & {\bf 40.9} & \bf 50.00 & \bf 31.86 & \bf 46.47 \\
		\midrule
		VSLNet* & 43.08 & 22.52 & 41.52 & 52.86 & 34.87 &  48.23 \\	
		VSLNet*-Q & 44.86 & 25.30 & 42.91 & 53.10 & 33.78 & 49.13 \\
		VSLNet*-V & 45.16 & 25.48 & 44.04 & {\bf 55.89} & {\bf 36.57} &  {\bf 51.60}\\
		BSSARD-VSLNet* & {\bf 47.20} & {\bf 27.17} & {\bf 44.59} & 55.65 & 36.33 & 50.45 \\
		\bottomrule
	\end{tabular}
	\caption{Ablation study about bias generator.}
	\label{backbones_charades_cd}
\end{table}

\begin{table}[t]
	\centering
	\renewcommand{\arraystretch}{1}
	\begin{tabular}{c c c c c}
		\toprule
		visual & language & r1i5 & r1i7 & mIoU\\
		\midrule
		before & before & {\bf 47.32} & 26.46 & 44.30 \\
		before & after & 47.20 & {\bf 27.17} & {\bf 44.59} \\
		after & before & 45.45 & 25.30 & 42.66 \\
		after & after & 46.67 & 25.90 & 43.69 \\
		\bottomrule
	\end{tabular}
	\caption{Ablation study about bias injection positions. ``before'' and ``after'' represent the injection position before and after the feature encoder, respectively.}
	\label{bias_injection_positions}
\end{table}

\subsubsection{Bias Generator}

We analyze the impact of visual and query bias generators on different backbones and show the results in Tables~\ref{backbones_charades_cd}.
*-V and *-Q refer to the backbone model that uses only visual and query bias generators, respectively. The results show that the visual and query bias generators are effective on all backbones, and the two generators complement each other. 
Besides, we find that the performance of VSLNet*-V is even better than BSSARD-VSLNet* on the IID test set. This is mainly due to two reasons, one is that the powerful cross-modality alignment ability of VSLNet* weakens the integration effect of the two bias generators, and the other is the presence of numerous language bias samples in the IID test set.

\subsubsection{Bias Injection Positions}

We explore the effect of bias feature injection positions. 
The injection position determines which features the bias generator needs to generate bias on and which parts of the grounding model need to complete debias task. 
For visual and language biases, we separately test the injection position before and after the feature encoder. The results are shown in Table~\ref{bias_injection_positions}. 
We can see that injecting the visual bias feature before the visual feature encoder and injecting the language bias feature after the language feature encoder is the most effective. 
This primarily depends on the distinct characteristics of vision and text data and the need for both accuracy and diversity in generated bias features. First, the intricate nature of video content
and the high-dimensional visual feature space pose serious challenges for the feature encoder that aims to map visual and textual features into a shared space for multi-modal feature alignment. Injecting visual bias before the feature encoder reduces the encoder’s learning complexity, enabling the model to differentiate real from synthesized visual features. Second, introducing language bias after the feature encoder helps the model better capture bias during the multimodal feature alignment stage. Third, multiple injection positions increase the diversity of generated bias features. 

Besides, we inject bias in the feature domain instead of the input data because it is more controllable to generate high-quality sample features. If we directly generate video samples, the generator should not only capture the bias information but also learn the original video representation, which greatly increases the learning difficulty. 

\begin{table}[t]
	\centering
	\renewcommand{\arraystretch}{1}
	\begin{tabular}{c c c c c}
		\toprule
		$z_p$ & concat & r1i5 & r1i7 & mIoU\\
		\midrule
		$\mathbb{R}^{n}$ & $F_T,F_V$ & 46.40 & 26.40 & 44.07 \\
		
		$\mathbb{R}^{3 \times n}$ & $F_T,F_V$ & {\bf 47.20} & {\bf 27.17} & {\bf 44.59}\\
		
		$\mathbb{R}^{3 \times n}$ & $F_T,F_V,F_S$ & 46.70 & 26.28 & 44.15 \\
		
		\bottomrule
	\end{tabular}
	\caption{Ablation study about fusion method of $z_p$.}
	\label{fusion method}
\end{table}

\begin{table}
	\centering
	\renewcommand{\arraystretch}{1}
	\setlength{\tabcolsep}{2mm}
	\begin{tabular}{c c c c}
		\toprule
		Training strategy & r1i5 & r1i7 & mIoU \\
		\midrule
		alternate each epoch & 46.79 & 26.76 & 44.53 \\
		
		random each step & 45.81 & 25.84 & 43.81 \\
		
		alternate each step & {\bf 47.20} & {\bf 27.17} & {\bf 44.59}\\
		
		\bottomrule
	\end{tabular}
	\caption{Ablation study about training strategy.}
	\label{Training strategy}
\end{table}

\subsubsection{Fusion Method of $z_p$ in VBG}
We investigate the impact of different fusion methods between $z_p$ and visual features in VBG. 
Table~\ref{fusion method} summarizes the results on the OOD test set.
We test three fusion methods, 
(1) encoding the temporal position as $z_p \in \mathbb{R}^{n}$ and fusing it with the temporal stream $F_T$ and video stream $F_V$.
(2) encoding the temporal position as $z_p \in \mathbb{R}^{3 \times n}$ and fusing it with $F_T$ and $F_V$.
(3) encoding the temporal position as $z_p \in \mathbb{R}^{3 \times n}$ and fusing it with the $F_T$, $F_V$ and spatial stream $F_S$. 
The results demonstrate that the second method is the most effective.
This is because $z_p \in \mathbb{R}^{3 \times n}$ can better represent temporal position information, and only the time stream $F_T$ and video stream $F_V$ containing temporal information that can better utilize temporal position information.

\begin{figure*}[t]
	\centering
	\subfigure[``cook'' suggests the target moment starts at the video start.]{
		\includegraphics[scale=0.3]{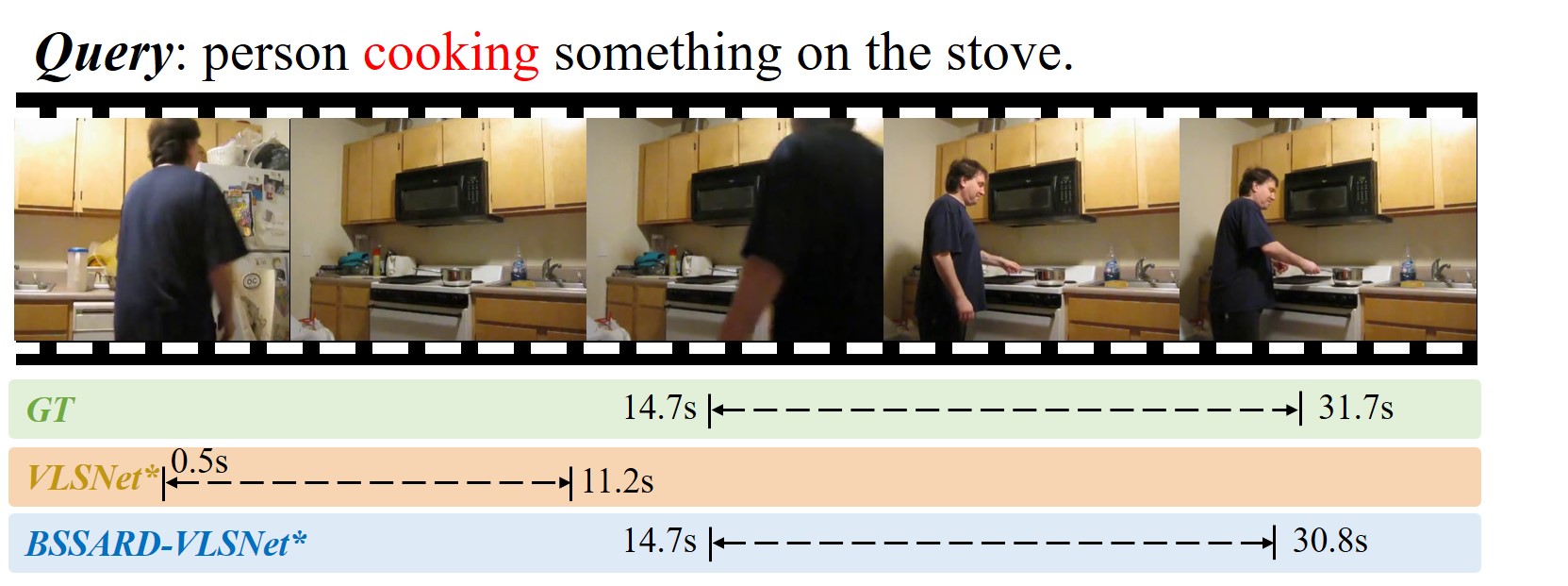}
		\label{visualization_sample1}
	}
	\quad
	\subfigure[``start'' implies the target moment is at the end of the video.]{
		\includegraphics[scale=0.3]{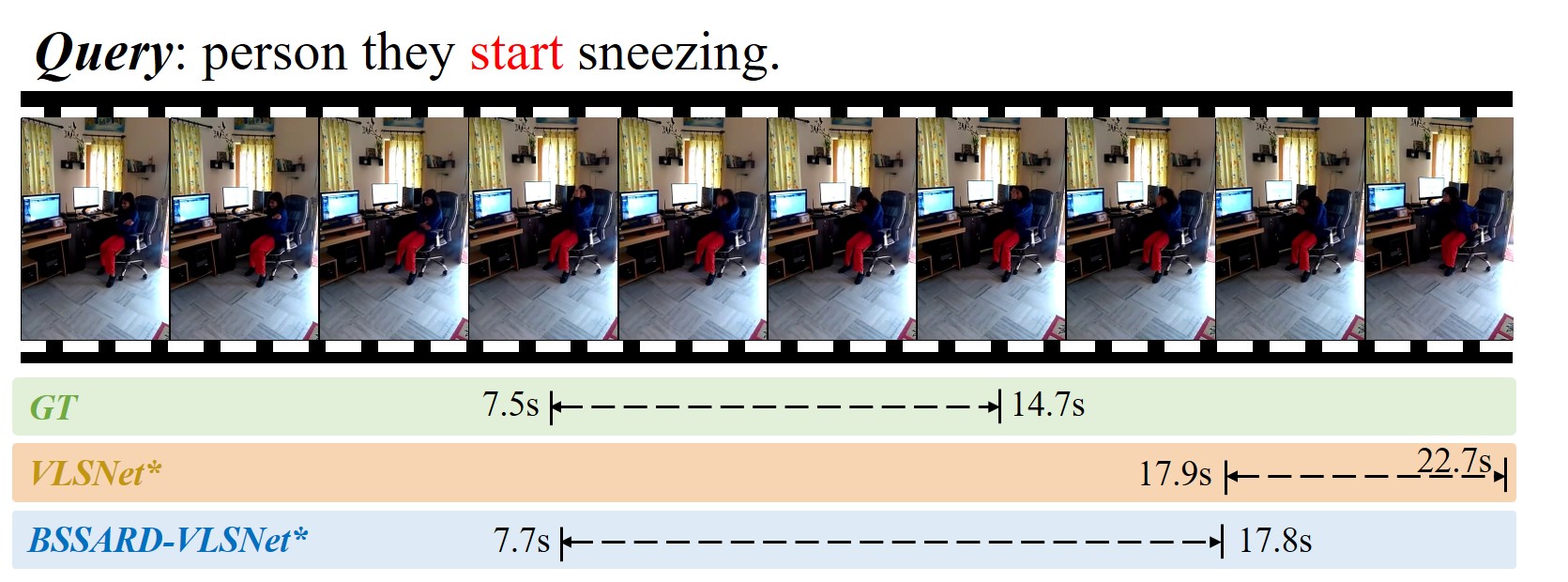}
		\label{visualization_sample2}
	}
	\quad
	\subfigure[``hold'' suggests the target moment starts at the video start.]{
		\includegraphics[scale=0.3]{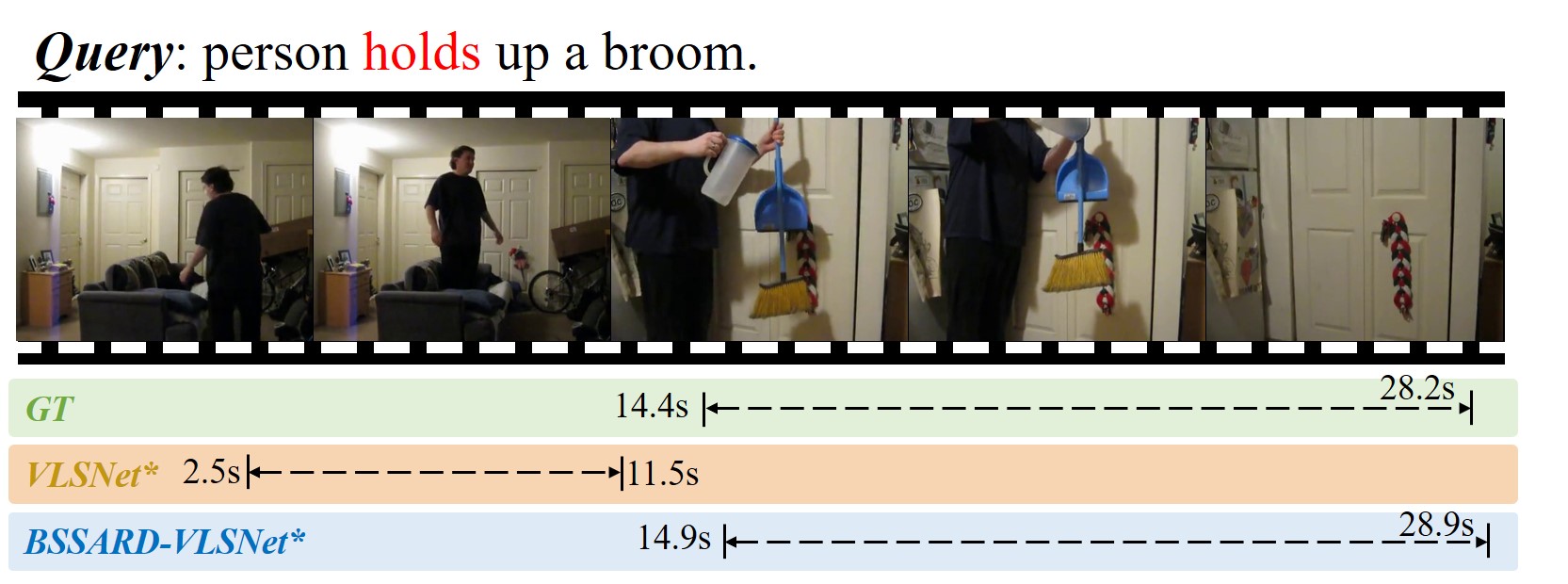}
		\label{visualization_sample3}
	}
	\quad
	\subfigure[``hold'' suggests the target moment starts at the video start.]{
		\includegraphics[scale=0.3]{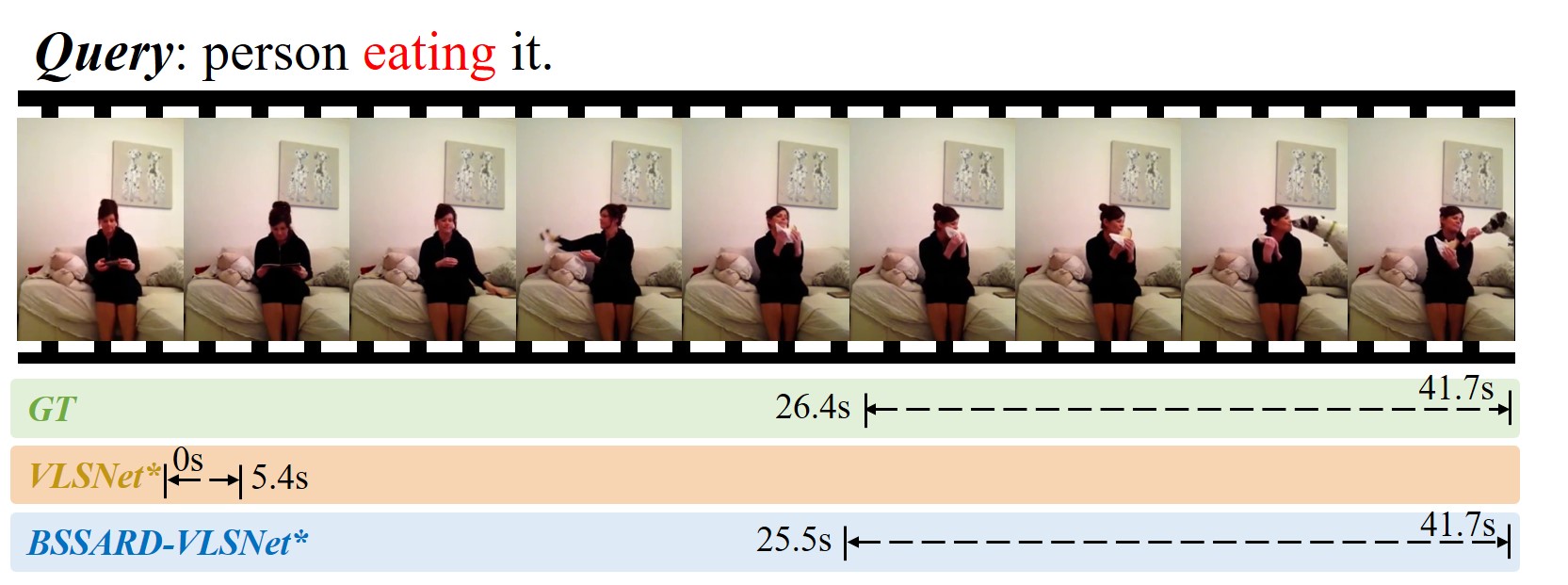}
		\label{visualization_sample4}
	}
	\quad
	\centering
	\caption{The visualization comparison results between VLSNet* and BSSAR-VLSNet*.}
	\label{visualization_sample}
\end{figure*}

\subsubsection{Training Strategy for Bias Generators}
We investigate the impact of the training order of two bias generators and show results in Table~\ref{Training strategy}.
We test three training strategies: (1) alternately training the two bias generators at each epoch. (2) randomly selecting one bias generator for training at each training step. (3) alternately training the two bias generators at each training step. 
The experimental results indicate that the third training strategy is optimal. This is mainly because alternate training the two bias generators can avoid the model overfitting into a certain debias approach.

\subsection{Analysis of the Debias Effect}

\subsubsection{The Verification of Mitigating Bias} 

We randomly shuffle the words in each query text, which disrupts potential spurious correlations among query text and target moments. It also prevents the model from relying on visual-language alignment unless it depends on visual bias. Table~\ref{ablation_removing_language_bias} shows BSSARD achieves more significant performance degradation. This indicates our approach relies less on bias correlations beyond query text to perform grounding.

\begin{table}[t]
	\centering
	\small
	\renewcommand{\arraystretch}{1}
	\setlength{\tabcolsep}{1.2mm}
	\begin{tabular}{c c c c c }
		\toprule
		Setting & B-VSLNet & VSLNet & B-QAVE & QAVE \\
		\midrule
		Original query & 47.20 & 43.08 & 44.47 & 37.84 \\
		Random query & 15.11 & 18.31 & 29.87 & 31.17 \\
		\midrule
		Gap & \textbf{67.99\% $\downarrow$ } & \textbf{57.50\% $\downarrow$} & \textbf{32.83\% $\downarrow$} & \textbf{17.63\% $\downarrow$} \\
		\bottomrule
	\end{tabular}
	\caption{Comparison results under random text input.}
	\label{ablation_removing_language_bias}
\end{table}

\subsubsection{Qualitative Analysis}

\begin{figure}[t]
	\centering
	{\includegraphics[width=1\linewidth]{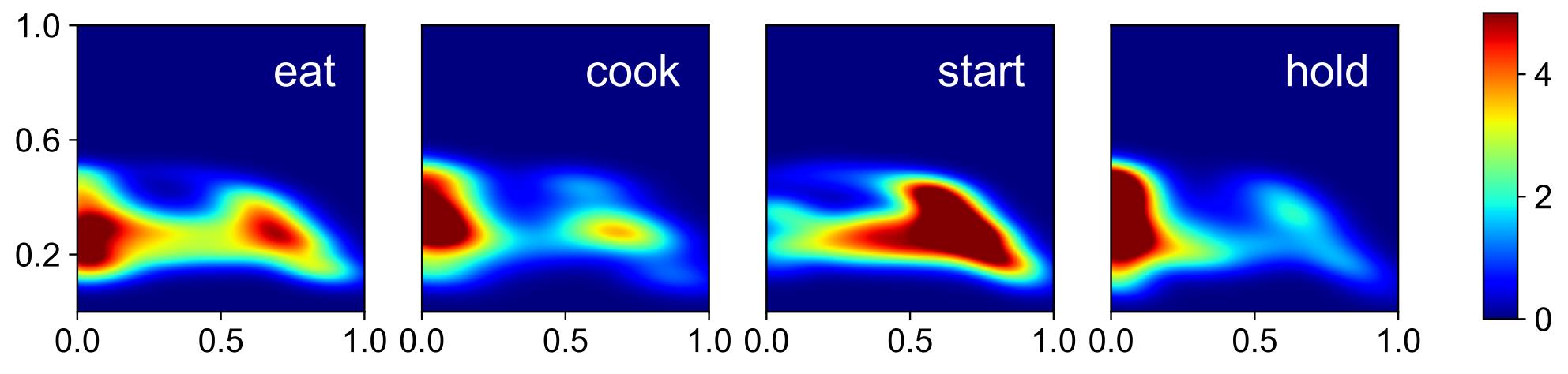}}
	\caption{The temporal distribution of target moments for video-text samples with certain common verbs. The horizontal and vertical axes denote the normalized starting time and duration of the target moment, respectively. The color represents the sample density.}
	\label{sample_bias_distribution}
\end{figure}

We also compare the debias effect of our BSSARD-VLSNet* and VLSNet* at the sample level in Figure~\ref{visualization_sample}. 
First, we give the temporal distribution of the target moment for samples with common verbs in the Charades-CD training set in Figure~\ref{sample_bias_distribution}. This indicates each verb contains prior knowledge for the localization of the target video clip. 
For example, the verb ``cook'' indicates that the temporal location of the target moments for most samples is the first half of the video. 
Hence, if the query text contains ``cook'' and the model will output the target moment with high probability in the first half of the video, which is the phenomenon of VLSNet* shown in Figure~\ref{visualization_sample1}. However, the result of our BSSARD-VLSNet* is not affected by the bias in the training set. 
The other three examples also show that the VLSNet* model relies on the bias in the training set for prediction, while our BSSARD-VLSNet* can effectively reduce the dependence on the bias and shift the model's attention back to cross-modality matching to make correct predictions.

